\documentclass[a4paper,fleqn]{cas-dc}

\usepackage[numbers]{natbib}
\usepackage[ruled,vlined]{algorithm2e}
\usepackage{hyperref}

\usepackage{xcolor}

\def\tsc#1{\csdef{#1}{\textsc{\lowercase{#1}}\xspace}}
\tsc{WGM}
\tsc{QE}
\tsc{EP}
\tsc{PMS}
\tsc{BEC}
\tsc{DE}
\begin{document}
\let\WriteBookmarks\relax
\def\floatpagepagefraction{1}
\def\textpagefraction{.001}
\shorttitle{Face to BMI prediction using Face Semantic Segmentation}
\shortauthors{Nadeem et~al.}

\title [mode = title]{Estimation of BMI from Facial Images using Semantic Segmentation based Region-Aware Pooling}

\author[1]{Nadeem Yousaf}
%\cormark[1]
%\fnmark[1]
%\ead{nadeem.yousaf@itu.edu.pk}

\credit{conceptualization, software, writing - original draft, writing-review \& editing}

\address[1]{Intelligent Machine Lab, Information Technology University, Pakistan}

\author[2]{Sarfaraz Hussein}
%\fnmark[2]
%\ead{cvr3@sayahna.org}
%\ead[URL]{www.sayahna.org}

\credit{ conceptualization, writing-review and editing, idea, formal analysis, supervision, project administration.}

\address[2]{Machine Learning and Data Science @ The Home Depot, USA}

\author%
[1]
{Waqas Sultani \corref{cor1}}
%\cormark[2]
%\fnmark[1,3]
%\ead{rishi@stmdocs.in}
%\ead[URL]{waqas.sultani@itu.edu.pk}

\credit{conceptualization, writing-review and editing, idea, formal analysis, supervision, project administration.}

%\address[1]{Information Technology University, Pakistan}

\cortext[cor1]{Corresponding author}
%\cortext[cor2]{Principal corresponding author}

\begin{abstract}
Body-Mass-Index (BMI) conveys important information about one's life such as
health and socio-economic conditions. Large-scale automatic estimation of BMIs can help predict several societal behaviors such as health, job opportunities, friendships, and popularity. \textcolor{black}{ The recent works have either employed hand-crafted geometrical face features or face-level deep convolutional neural network features for face to BMI prediction.  The hand-crafted geometrical face feature lack generalizability and face-level deep features don’t have detailed local information.} Although useful, these methods missed the detailed local information which is essential for exact BMI prediction. In this paper, we propose to use deep features that are pooled from different face regions (eye, nose, eyebrow, lips, etc.,) and demonstrate that this explicit pooling from face regions can significantly boost the performance of BMI prediction. To address the problem of accurate and pixel-level face regions localization, we propose to use face semantic segmentation in our framework. Extensive experiments are performed using different Convolutional Neural Network (CNN) backbones including FaceNet and VGG-face on three publicly available datasets: VisualBMI, Bollywood and VIP attributes.   \textcolor{black}{Experimental results demonstrate that, as compared to the recent works, the proposed Reg-GAP gives a percentage improvement of 22.4\% on VIP-attribute, 3.3\% on VisualBMI, and 63.09\% on the Bollywood dataset.}
\end{abstract}


\maketitle

\section{Introduction}

Faces depict important information about one’s personality e.g., age, gender, race, psychological conditions, poverty level as well as health conditions. To measure the overall health condition of a person, usually, body weight and height are used which are encoded in Body-Mass-Index (BMI). Person's BMI has a significant impact on several aspects of life, including health \cite{d1,d2,d3,d4}, job opportunities \cite{Fat_chance}, friendships and popularity \cite{Homophily}. Recent studies demonstrate that higher BMI can lead to many diseases such as heart disease \cite{d2,d3}, diabetes \cite{d1,d2,d4}, stroke \cite{d3}, cancer \cite{d1,d2,d3,d4}, sleep apnea \cite{d1,d3,d4}, hypertension \cite{d1,d3}, fatty liver disease \cite{d4}, kidney disease \cite{d4}, depression \cite{d4}, and pregnancy problems \cite{d5}. Other than health, BMI values have also been used to estimate and predict the social behaviors of societies. For example, on social media, people with similar BMI values are more likely to make connections as compared to people with dissimilar BMI values \cite{Homophily}. Similarly, people with higher BMI values have fewer followers as compared to people with lower BMI values \cite{Homophily}. Similarly, Caliendo et al \cite{Fat_chance} pointed out that obese women (higher BMI) observe weight-based discrimination during job interviews. Consequently, obese women find it more difficult to find a job and they get low wages as compared to their less obese counterparts. 

\begin{figure*}
\centering
\includegraphics[width=\linewidth]{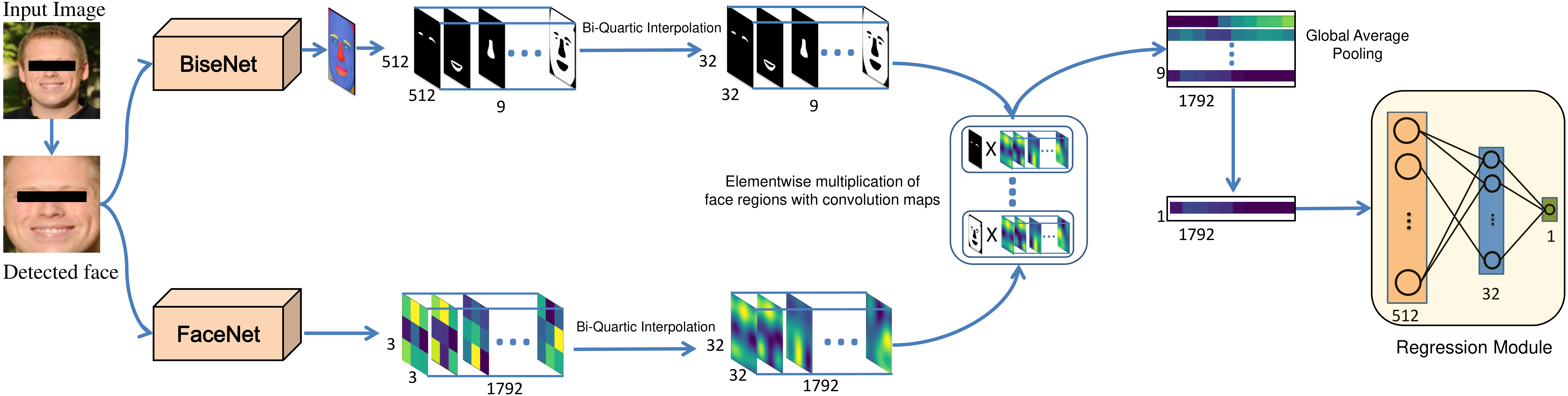}

\caption{The pipeline of the proposed approach. Given the input image, we crop the face region employing face detection \cite{mtcnn}. Each face region mask is obtained through face semantic segmentation. After that, face region masks are element-wise multiplied with the convolution feature maps to give high weights to different face regions. Global average pooling is then applied to each masked convolution map separately. Finally, we employ the regression module to obtain the BMI prediction.}
\label{fig:Blockdiagram}
\end{figure*}

In the past, researchers have studied the association between facial measures and body weight. It has been observed that BMI is strongly correlated with eye-detailed information (e.g., intraocular pressure (IOP) and anterior corneal curvature (ACD) \cite{Panon_Corre}, neck circumference \cite{reg1,reg2} and face physical measures such as width-to-height ratio, perimeter-to-area ratio, and cheek-to-jaw-width ratio \cite{reg4}. However, most of these works have used hand-crafted features and took face measurements manually. Due to the large impact of BMI on a person's life and society behaviors, it is of significant importance to measure the BMI on a large scale. However, this would need a lot of resources to go door-to-door to compute each individual's BMI. Fortunately, computer vision and deep learning provide a non-intrusive, efficient, and cheaper way to estimate people's BMI through their face images on social media. \cite{EnesKocabey, Homophily}

Although it has been well-established \cite{Panon_Corre,reg5} that different facial regions are strongly correlated with BMI values and their measurements can help better prediction of BMI, all the previous deep learning-based methods have used the deep features from the full face images. For example, Pascalia et al., \cite{CBM_face} proposed a method for automatic extraction of geometric features using 3D facial data acquired with low-cost depth scanners. They have experimentally shown that these features are highly correlated with weight and BMI. Similarly, authors in \cite{CBM_fct} employed diffusion tensor imaging on white matter alterations to estimate obesity and BMI. We, on the other hand, explicitly use the information from different facial regions to obtain a robust feature embedding. To exploit the relationship of different facial regions with BMI, we obtain the improved feature vector by pooling the convolution feature maps based on different semantic regions of the face. We obtain different face regions through face semantic segmentation. The face semantic segmentation provides accurate pixel-wise locations of different face regions. Figure \ref{fig:Blockdiagram} demonstrates the pipeline of the proposed approach. 
Assuming FaceNet \cite{facenet} as a face feature extractor module, given the input image, we crop the face region employing face detection \cite{mtcnn}.  Face region masks of different face regions such as ear, eyes, eyebrow, hair, lips, neck, nose, and skin are obtained through face semantic segmentation. After resizing face-region masks using bi-quartic interpolation, each mask is multiplied with convolution feature maps to obtain feature map values, specific to different face regions. After that, we perform \textbf{Reg}ion aware \textbf{G}lobal \textbf{A}verage \textbf{P}ooling (Reg-GAP) to get the final embedding for training BMI prediction regression module. Finally, the regression module is employed to predict the BMI value of the face. Although simple and straightforward, our experimental results demonstrate that face semantic segmentation-based feature pooling helps improve BMI prediction on two popular publicly available datasets.

The overall organization of the paper is as follows: Section 3 summarizes some of the recent research works for BMI prediction, section 4 provides details of the proposed methodology, section 5 shows experimental results and analysis and finally, section 6 concludes the paper.

\section{Related work}

Due to the significant impact of BMI on health, economic conditions, societal behaviors, several researchers have extensively worked to accurately estimate BMI from people's eyes and neck information, facial dimensions, and face and human body images.

Recent studies have shown the facial features encode useful information about a person's health. Panon et al. \cite{Panon_Corre} investigated the relationship between body mass index and ocular parameters. Employing enhanced depth-imaging optical coherence tomography, several measurements of anterior and posterior segment parameters of the eye measurements were made. The segments include anterior chamber angle, central corneal thickness, macular thickness (MT), anterior chamber depth (ACD), ganglion cell thickness (GCT), retinal nerve fiber layer thickness among many others. Moreover, anterior corneal curvature and intraocular pressure (IOP) were measured by non-contact tonometry. Using data from fifty-three left eyes of normal weight subjects and 67 age-sex matched overweight subjects, they concluded that intraocular pressure (IOP) and anterior chamber depth (ACD) are positively correlated with BMI. 

Coetzee et al. \cite{reg4} demonstrated that there are two important prerequisites for any health cue. One of them is the perception of weight in the face which can significantly predict perceived health and attractiveness. Authors in \cite{reg5} tried to spot the facial cues that are associated with BMI. They recruited two groups of African and two groups of Caucasian participants, determined their BMI, and measured their 2-D facial images for perimeter-to-area ratio, cheek-to-jaw-width ratio, and width-to-height ratio, The width-to-height, and cheek-to-jaw-width ratios were found to be significantly associated with BMI in males and females. Mayer et al. \cite{reg6} studied to assess the association of BMI and waist-to-hip ratio (WHR) with facial shape and texture in females. The females included in the study were middle-aged European women with a BMI between 17-35. They showed that BMI is better predictable than WHR from facial attributes. Saka et al. \cite{reg1} performed a pilot study on Turkish adults and found that neck circumference can be utilized as an indicator for abdominal obesity and similarly, Atwa et al. \cite{reg2} have utilized neck circumference to detect children with high BMI. %\ny{The authors of \cite{CBM_fct} developed a spatially guided enhanced FCT (s-eFCT). They computed and used the regional fractional anisotropy (FA) and the mean diffusivity (MD) values to predict BMI values and found the correlation between real and predicted BMIs of 0.57.}

Segmentation has been used in several medical image analyses to improve detection and classification accuracy. Qayyum et al., \cite{CBM_seg1} proposed a hybrid 3D residual network (RN) with a squeeze and excitation (SE) block for volumetric segmentation of kidney, liver, and their associated tumors. %They used Kidney Tumor Segmentation 2019 dataset and the public MICCAI 2017 Liver Tumor Segmentation dataset for their experiments. 
The authors in \cite{CBM_seg2} proposed the NucleiSegNet - a robust deep learning network architecture for the nuclei segmentation of hematoxylin and eosin-stained liver cancer histopathology images. Their proposed deep-learning architecture yielded superior results compared to state-of-the-art nuclei segmentation methods.
 Similarly, Mussi et al., \cite{CBM_ear} proposed an algorithm that performs ear depth map segmentation.
 %into main elements with the image. It was tested on the dataset of 150 ear images.

The first work to show that the BMI of a person can be automatically predicted from a 2D face image using the geometrical features was done by Wen and Guo \cite{wen_IVC_2013}. After detecting the face and eyes, they normalized the face based on the eyes' coordinates. Normalization was done to align the face images into common eye coordinates. Next, the active shape model (ASM) was used to detect several key points in each face image. Seven geometrical features were detected which include width to upper facial height ratio, cheekbone to jaw width, eye size, perimeter to area ratio, lower face to face height ratio, mean of eyebrow height, and face width to lower face height ratio. They normalize these features before applying support vector regression (SVR). They evaluated the method on 14, 500 images from the MORPH-II dataset which is not freely publicly available. Following \cite{wen_IVC_2013}, Jiang et al. \cite{Body2BMI} extracted geometric features from the whole body to predict BMI from whole-body images.  \textcolor{black}{In contrast to whole-body images, face images are more easily available (on National ID cards, driving licenses, etc) and usually have little or no occlusion. Furthermore, to the best of our knowledge, there does not exist any publicly available body to BMI dataset}

With the resurgence of deep convolution neural networks, several interesting and new problems, including BMI prediction, have been efficiently addressed with improved accuracy. The first work related to deep learning-based BMI estimation from face photos was done by Enes Kocabey et al. \cite{EnesKocabey}. Instead of using traditional machine learning to extract hand-crafted features, they employed pre-trained deep neural network models to extract features. They have used two models to extract the features. The first model named `VGG-Face' \cite{vggface} was trained for the face recognition task and the second model named `VGG-Net' \cite{vggnet} was trained for general image classification. The features were extracted from the fully connected ($f_{c_6}$) layer. To perform the prediction, the epsilon support vector regression model \citep{svR} was employed. Furthermore, they have collected their dataset using Reddit-subreddit called progress-pics. Dantcheva et al. \cite{showme} proposed a CNN-based method to estimate the height, weight, and BMI using 50-layers ResNet-architecture. They had also presented a new `VIP-Attribute' dataset consisting of 1026 subjects. This dataset contains 513 males and 513 females.

Similarly, Jiang et al. \citep{labeldist} introduced a label distribution-based method for BMI estimation from face images. Their proposed approach contains two stages. In the first stage, BMI-related features are computed, and in the second stage, a label distribution-based BMI estimator is learned. Specifically, in the first stage, they utilized a face model \cite{wen2016} which was originally trained for the face recognition task. They used the FIW-BMI dataset \cite{FIW_dataset} to fine-tuned the face-recogn-ition model \cite{wen2016} to a BMI-related face model by replacing the last fully connected layer of 512 dimensions into 1 and use the Euclidean loss. They defined a single BMI value as a discrete probability distribution over the range of BMIs using Gaussian distribution and triangle distribution. After extracting the facial features, five estimators were learned. The estimator includes: Principal Component Analysis (PCA), Support Vector Regression (SVR), Gaussian Process Regression (GPR), Partial Least Square analysis (PLS), Canonical Correlation Analysis (CCA), and two label distribution (LD) based estimator include (LD-CCA, LD-PLS). \textcolor{black}{Authors in [3,9,33] are using deep learning-based models as black-box i.e., they entirely rely on the network to extract meaningful features to map the face image to the BMI score. On the other hand, our proposed approach explicitly extracts sub-regions of the face and uses these local cues to learn an attention-based feature-space which is a more meaningful representation.}

Recently, human face and body semantic segmentation have been used in several computer vision applications. Khalil et al. \cite{semantic1} have used semantic face segmentation for gender and expression analysis. They segmented the facial images into six semantic classes: hair, skin, nose, eyes, mouth, and back-ground using a random decision forest. In their final step, they trained a Support Vector Machine (SVM) classifier for gender using the corresponding probability maps of facial regions. Improved facial attribute prediction based on face semantic segmentation was presented by \cite{Mahdi_2017_CVPR}. The core idea of their research was that facial attributes describe local properties and the probability of an attribute to appear in a face image is far from being uniform in the spatial domain. They obtained an improved facial attributes prediction while using localization cues from facial semantic segmentation. Similarly, Kalayeh et al. \cite{Mahdi_2018_CVPR} presented an approach for improved person re-identification using human semantic parsing.

Attention-based networks have shown promising results in several vision tasks. Wang et al. \cite{attention1} put forwarded an attention-based multi-branch network for makeup-invariant face verification. Authors in \cite{mva_reg_age} used the region-wise modelling to predict the human facial skin age and \cite{mva_reg_grow_cancer} used the multi-step region growing for segmentation of skin cancer images. Similarly, researchers in \cite{mva_fac_exp_rec} employed facial regions geometric features for the analysis of in and out-group differences in Western and East Asian facial expression recognition. \textcolor{black}{Similar to us, researchers in \cite{RAP} have also used  Region-based Average Pooling for context-aware object detection. Our proposed approach has several
differences from the approach presented in \cite{RAP}. The main purpose of RAP in \cite{RAP} is to combine the features of different object regions to achieve improved \textit{object} detection. However, on the other hand, we employ region-aware pooling to explicitly pool features from different face regions to make face to BMI prediction better. The approach in \cite{RAP} aims at learning the relationship between different regions and use these relationships to better classify and regress each region. Each region in \cite{RAP} corresponds to a single complete object. That is why they un-pooled the averaged vectors and concatenated them with original representations of each region (object) separately. In our case, each region is a sub-part of one complete object (face) and our purpose of averaging the regions and then combining them is to learn the individual importance of each region for the face to BMI estimation. Therefore,  we directly pass the averaged representation to the regression module instead of unpooling it and concatenating it to the region's representations. Furthermore, in \cite{RAP}, since each region is a different object, thus combining features of unrelated classes, such as a car and a cat, might not always result in optimal performance. While in our case, since the regions are a part of the same object, combining them will always complement the learning process.}

In contrast to the above-mentioned research works, in this work, we propose to use face semantic segmentation to extract local facial regions. Our proposed facial regions-based pooling provides robust feature embedding for face to BMI prediction. To the best of our knowledge, we are the first ones to propose facial regions-based pooling for BMI prediction.

\section{Methodology}

The proposed approach for BMI prediction from face images contains three main components. Given the face image, we extract deep features and employ semantic segmentation to obtain pixel-level localization of different face regions. After that, we integrate the semantic segmentation to obtain face-regions based pooling from convolution layers of the neural network which was trained on face images. Finally, the pooled features are used to predict BMI values using a fully-connected regression module. \textcolor{black}{The complete algorithm of our approach is given in Algorithm 1.}
Below we describe each component of the proposed approach in detail.
\begin{algorithm}[]
\SetKwInput{KwInput}{\textcolor{black}{Input}}                % Set the Input
\SetKwInput{KwOutput}{\textcolor{black}{Output}}              % set the Output
\DontPrintSemicolon
  
  \KwInput{\textcolor{black}{People Images Set I}}
  \KwOutput{\textcolor{black}{Linear BMI Prediction $B_1...B_k$}}

% Set Function Names
  \SetKwFunction{FPro}{\textcolor{black}{Predict\_BMI}}

% Write Function with word ``Function''
  \SetKwProg{Fn}{\textcolor{black}{Procedure}}{:}{}
  \Fn{\FPro{\textcolor{black}{I}}}{
  \textcolor{black}{
                    $D_1...D_k \xleftarrow\; Face\_Detection(I)$ \\
                    $F_1...F_k \xleftarrow\; VGGFace(D_1...D_K) $ \\
                    $S_1...S_k \xleftarrow\; Face\_Semantic\_Segmentation(D_1...D_k)$ \\
                    $\hspace{.065in}\hspace{.1in}$ $S_i$ $\in \hspace{0.2cm}$  ${\rm I\!R}^{h\times w\times j}$ \\
                    $M_1...M_k \xleftarrow\; PreProcess\_Masks(S_1...S_k)$ 
                    }

\For{\textcolor{black}{k=1 to K}}    
        { 
        \For{\textcolor{black}{j=1 to J}}    
        { 
\textcolor{black}{
        	${R}_{kj}$ =  ${F}_{k}$  $\star$ ${M}_{kj}$ }
        	
        	End} 
        
        End}
        \textcolor{black}{
     $v_{f_1}...v_{f_K} \xleftarrow\; Reg\_GAP({R}_{kj}$) \\
    $B_1...B_K \xleftarrow\; Regression\_Module(v_{f_1}...v_{f_K}$)\\
    \Return $\hspace{0.2cm}$ $B_1...B_k$  \\
  End} }
  
   \caption{\textcolor{black}{Algorithm to Estimate BMI from Facial Images}}
\end{algorithm}
 
\subsection{Face Feature Extraction} 

We employ two face feature extraction models: FaceNet \cite{facenet} and VGG-face \cite{vggface}. Below, we briefly describe both of the feature extraction methods.\\

\begin{figure}[hbt!]
\centering
\includegraphics[width=\linewidth]{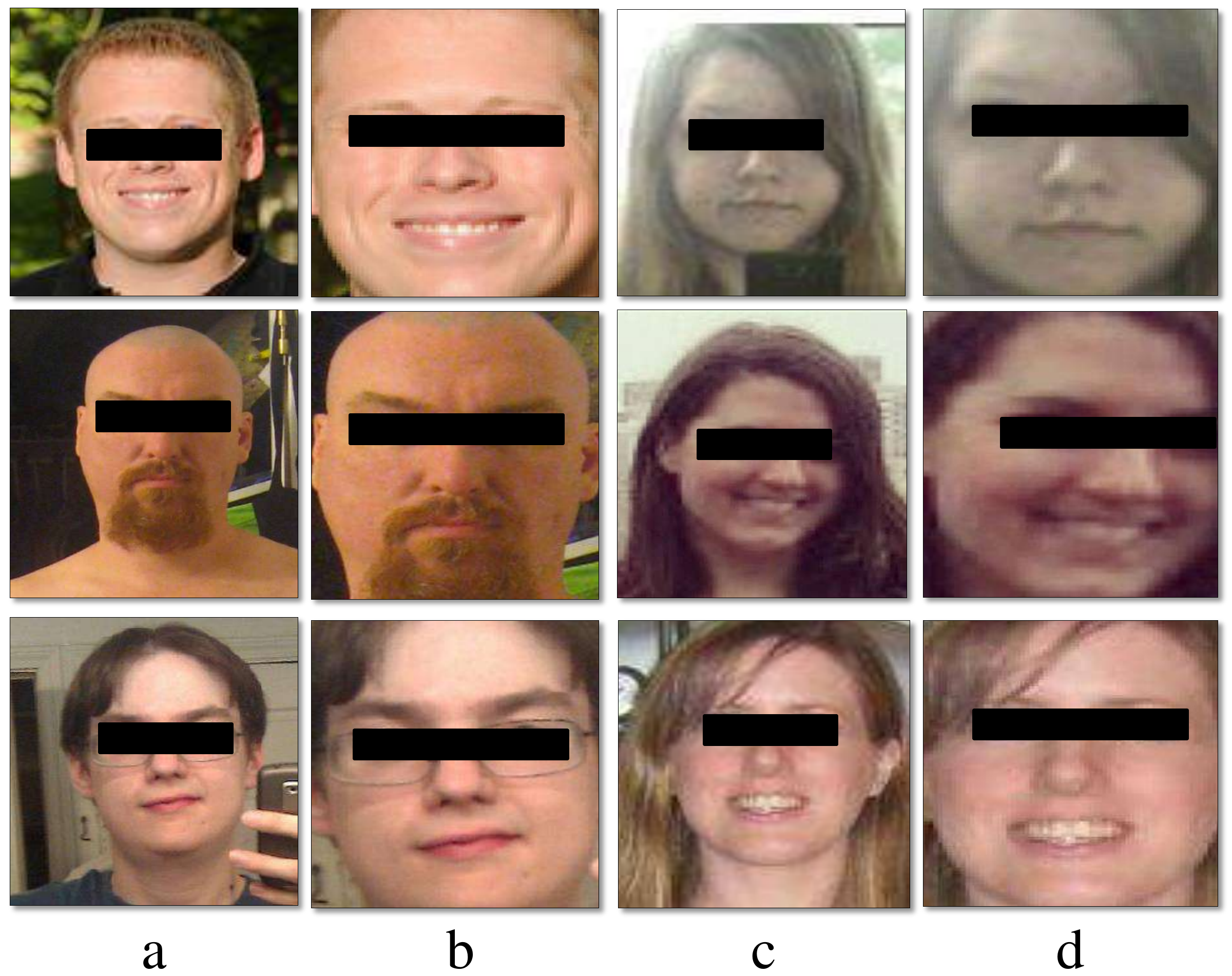}
\caption{Typical examples of face detection to extract FaceNet features are shown in this figure. The face detection is done using Multi-task Cascaded Convolutional Neural Networks \cite{mtcnn}. (a) and (c) shows the original face images and (b) and (d) shows the cropped images after face detection.}
\label{fig:FaceDeetction}
\end{figure}

\noindent\textbf{Feature extraction with FaceNet: } FaceNet directly learns a mapping from face images to a compact Euclidean space for tasks such as face recognition and verification. Similarly, face image clustering can be easily implemented using standard techniques by utilizing FaceNet embeddings as feature vectors. FaceNet model is trained on faces detected using multitask cascaded convolutional neural networks (MTCNN) \cite{mtcnn}. The input image shape required for FaceNet is of size 160$\times$160$\times$3. We first resize the images to this shape and then apply MTCNN. The MTCNN approach predicts face and landmark locations in a coarse-to-fine manner by adopting a cascaded structure with three stages of carefully designed deep convolutional networks. Figure \ref{fig:FaceDeetction} represents the detection of facing using MTCNN on VisualBMI dataset \cite{EnesKocabey}. In this work, we extract the features maps of size 3$\times$3$\times$ 1792 from % $Block$\textunderscore$8$\textunderscore $6$\textunderscore$ScaleSum$ 
last convolution layer of FaceNet, which was resized to 32 $\times$ 32 $\times$ 1792 using bi-quartic interpolation. Facial-regions feature extraction (see Section 4.3) is applied to these feature maps to extract the region aware features from the FaceNet model.\\

\noindent\textbf{Feature extraction with VGGFace: } The second face extraction deep convolution neural network that we use in our experiments is VGGFace \cite{vggface}.
VGGFace is trained as a face classifier with 2.6 million facial images of more than 2600 people. The name `VGGFace' was given later to the model and it was described by Omkar Parkhi in the 2015 paper titled “Deep Face Recognition \cite{vggface}. The input shape required for VGGFace is 224$\times$224$\times$3. We resize the images to this shape without applying face detection since this model was originally trained in these settings. We use the features from the $Conv5$\textunderscore$3$ layer of VGGFace which has a shape of 14$\times$14$\times$512. In the experiments, we resize the feature maps to 32$\times$32$\times$512 with the help of bi-quartic interpolation. Finally, facial-regions feature extraction (see section 4.3) is applied to these feature maps to extract the region aware features of the VGGFace model.

\subsection{Face semantic segmentation}
\begin{figure}[t]
\centering
\includegraphics[width=\linewidth]{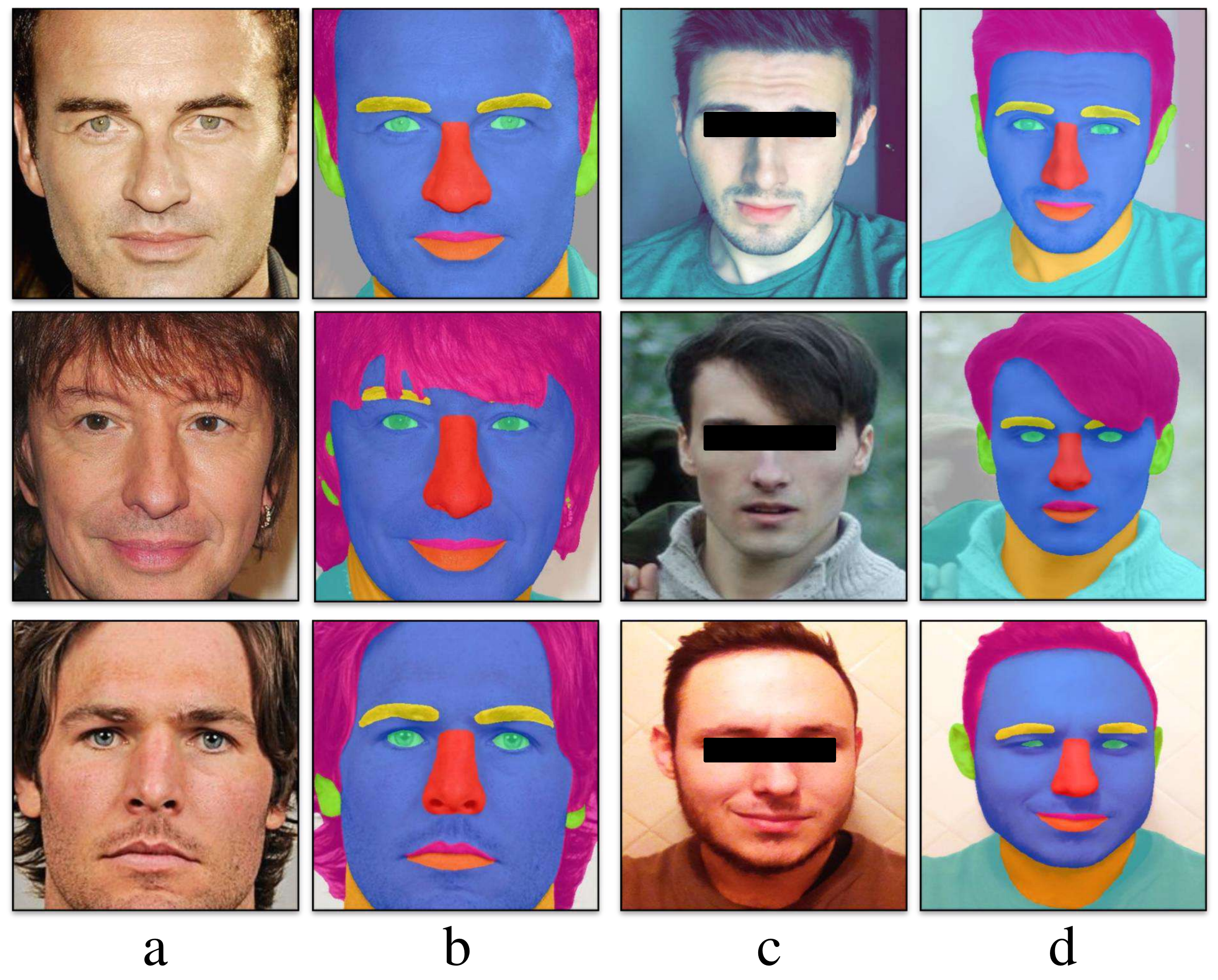}
\caption{Examples of the face semantic segmentation. The first and the third column show the face images  from the VIP attribute dataset and VisualBMI dataset while second and the fourth column shows their resultant face semantic segmentation.}
\label{fig:parsing}
\end{figure}

\begin{figure*}
\centering
\includegraphics[width=\linewidth]{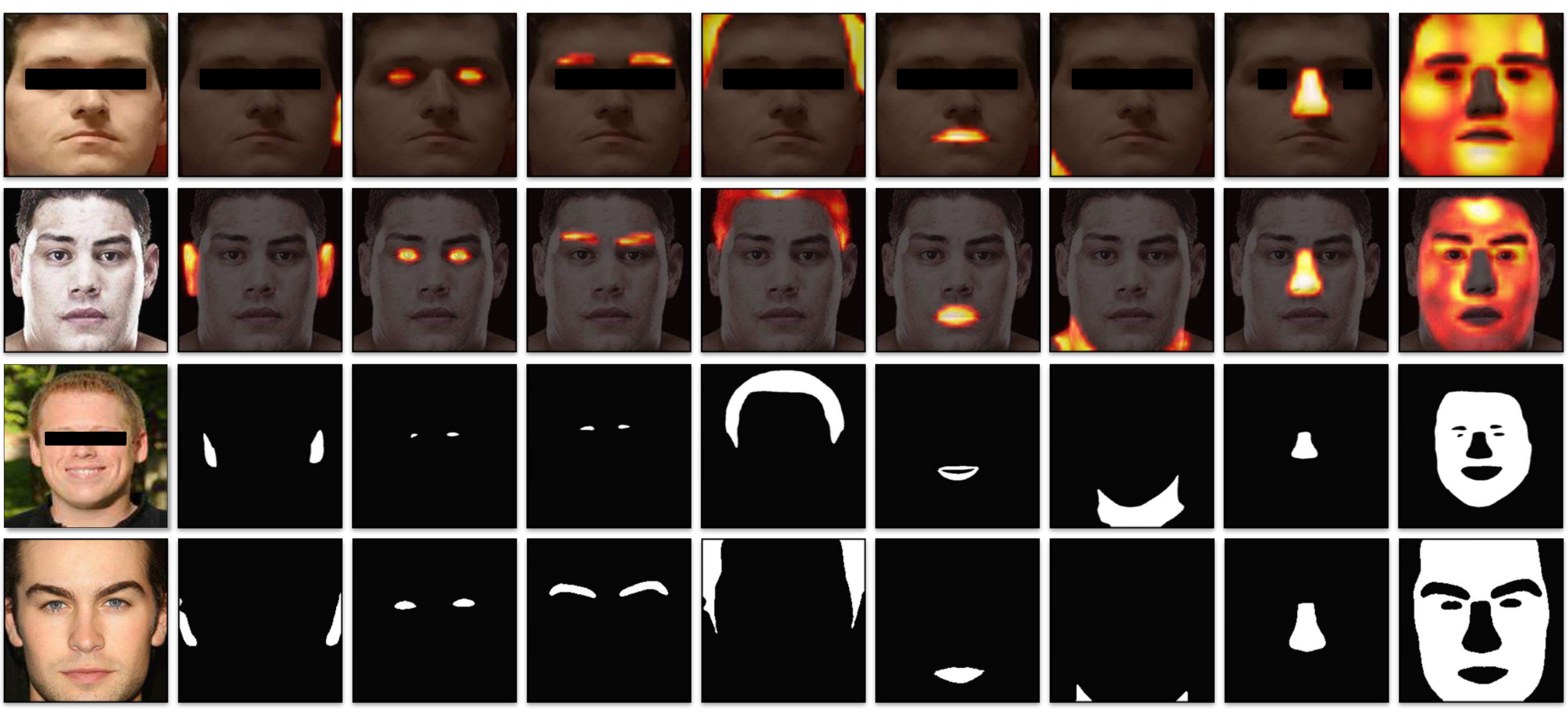}
\caption{Given the face semantic segmentation, we extract different face regions and explicitly pool features from those regions. The regions are (Left to Right): ear, eyes, eyebrow, hair, lips, neck, nose, and skin. The bottom two rows show the binary mask obtained from segmentation and the top two rows show region corresponding feature maps. The first and third row samples are from the VisualBMI dataset while the second and fourth-row samples are from the VIP attribute dataset.}
\label{fig:featuresmasks}
\end{figure*}

 To accurately localize different facial regions, we employ a bilateral segmentation network for face parsing \cite{bisenet}. To obtain real-time segmentation with sufficient accuracy, authors in \cite{bisenet} use small stride to preserve the spatial information and employs a fast down-sampling technique to obtain a sufficient receptive field. Specifically, to avoid losing spatial information due to small image size, a 3-layer convolution network is employed that output feature maps that are 1/8 of the original image. Similarly to preserve contextual information through a large receptive field, authors in \cite{bisenet} propose to employ the Xception network along with the global average pooling layer followed by U-structure to fuse the features.
 Finally, the network is trained end to end using softmax loss function which is given by 
\begin{eqnarray}
  \label{eq:softma_bisenet}
  loss = \frac{1}{N}\sum_{k}^{}-log\left ( \frac{e^{o_{k}}}{\sum_{j}e^{o_{j}}} \right ),
\end{eqnarray}
where o is the output of the network.
We encourage readers to this reference \cite{bisenet} for the model details of segmentation network architecture. The authors in \cite{bisenet} demonstrated the results on Cityscapes,
CamVid, and COCO-Stuff datasets. Since we are interested in face image segmentation, we use the model pre-trained on CelebAMask-HQ dataset \cite{CelebAMask-HQ}.\textcolor{black}{ We have used modified BiseNet which produces precise segmentation results as shown in Figure 3 and Figure 4. There are several differences between modified Bisnet and the originally proposed BiseNet such as 1) Original BiseNet take the image of the whole scene as input while in our modified version,  we first apply face detector (MTCNN) to localize the face and then input it to BiseNet which improves the accuracy for the face parsing, 2) As shown in \cite{bisenet}, original BiseNet was trained on two models: Xception and ResNet. Xception has fewer parameters which make it faster but had lower segmentation accuracy of 71.4\% as compared to ResNet which had more parameters and has better accuracy of 78.9\%. Therefore, our modified version is using ResNet as its backbone to ensure the highest accuracy as compared to its faster version with Xception as a backbone.} The typical examples of face semantic segmentation on VisualBMI \cite{EnesKocabey} and VIP attributes \cite{showme} datasets are shown in Figure \ref{fig:parsing}. We modified the BiseNet model to generate a separate binary mask for each region instead of a combined mask of all regions as shown in Figure \ref{fig:featuresmasks}. The separate binary masks are later pre-processed according to input image size and are shown in the bottom two rows of Figure \ref{fig:featuresmasks}.

\subsection{Region-aware Global Average Pooling}
We employ face semantic segmentation to pool the deep features from face regions. In Figure \ref{fig:featuresmasks}, in the bottom two rows, we show the masks obtained from different face regions. The face region includes the ear, eyes, eyebrow, hair, lips, neck, nose, and skin, and background. To obtain the face-region aware features, we perform element-wise multiplication of feature maps with that of mask obtained from semantic segmentation. Figure \ref{fig:featuresmasks}, in the top two rows, showed the region-aware features obtained after element-wise multiplication of masked regions with feature maps where feature maps are obtained using FaceNet or VGGFace.

\begin{figure}[h!]
\centering
\includegraphics[width=\linewidth]{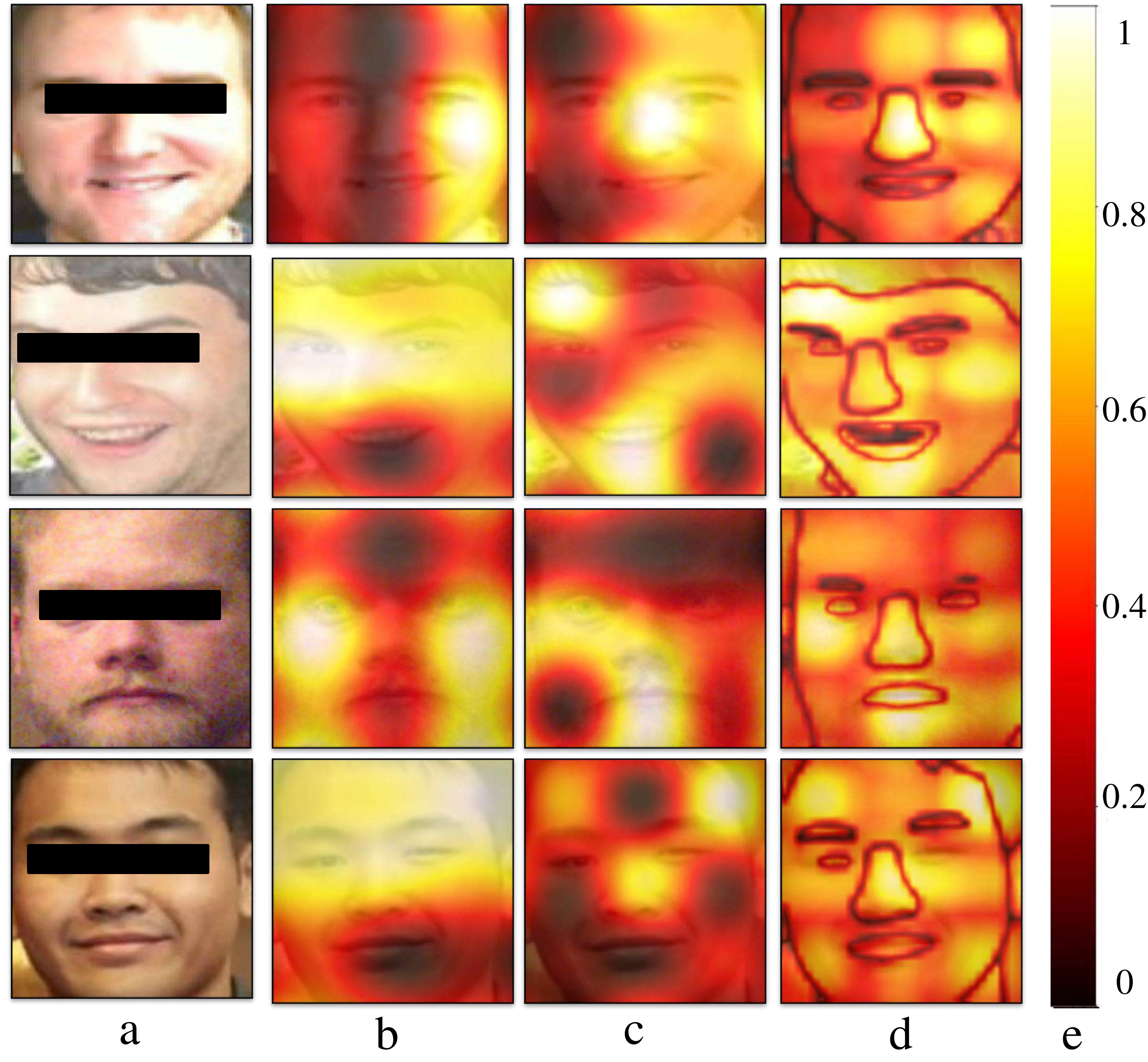}
\caption{This figure shows the comparison of convolution feature maps obtained by the global average pooling (GAP) and by proposed region-aware global averaging pooling (Reg-GAP). (a) shows the input to the FaceNet model, (b)is the mean across channels of the last convolution layer, (c) is the max across channels of last convolution layer, (d) is the max across the channels after Reg-GAP.}
\label{fig:featuresmaps}
\end{figure}
Formally, assume $\mathcal{F}$ represents convolution feature maps of FaceNet or VGG-Face and $\mathcal{M}_{k}$ represents stacked binary mask for region $i$ where $i$ can be eye, nose, necks, etc. 
\begin{eqnarray}
  \label{eq:regions_avg}
  \mathbf{r}_{i} = \frac{1}{N} \sum \sum \left ( \mathcal{F}  \star \mathcal{M}_{i}  \right),
\end{eqnarray}\label{eq:Maskedpool}where $\mathbf{r}_{i}$ is the global average pooled vector of face region $i$ and $\star$ represents Hadamard product. We repeat the steps mentioned in Eq 2 for each region separately. Finally, region aware global average pooled feature vector (Reg-GAP) is given by
\begin{eqnarray}
  \label{eq:reg_gap}
  \mathbf{r}_{Reg-Gap} = \frac{1}{K} \sum\left ( \mathbf{r}_{i} \right ),
\end{eqnarray}
where K is the number of regions.

In the experiments, we have also compared Reg-GAP with well known global average pooling (GAP) on the original convolutional feature maps. The GAP is defined as:
\begin{eqnarray}
  \label{eq:GAP}
 \mathbf{r}_{Gap} = \frac{1}{N} 
 \sum \sum\left ( \mathcal{F} \right ),
 %\sum_{i=1}^{N}\sum_{j=1}^{N}\left ( F_{ij} \right )
\end{eqnarray}where $\mathcal{F}$ is the original output of the last convolution layer of the model used.

In Figure \ref{fig:featuresmaps}, we show the comparison between the original FaceNet features maps and region-aware FaceNet features maps. The first column shows the cropped image which is input to the FaceNet model. The middle column shows the original feature maps extracted from  FaceNet and the last column show the region-aware feature maps. For illustration purposes, the shown feature maps (second column) are generated by taking mean across channels of original feature maps and we overlay it on the face image. The feature maps (third column) are generated by taking max across channels. Finally, to show the region-aware feature maps (last column), we took the max across the channel after taking element-wise multiplication with each face region. It can be seen that Reg-GAP feature maps capture more details of the face and are invariant to face variations.

%$Block$\textunderscore$8$\textunderscore $6$\textunderscore$ScaleSum$ layer of 

.\subsection{Regression module}
Once the region-aware feature vector is obtained, we employ the regression module to obtain the final Face to BMI prediction.  
The first layer of the architecture has 512 neurons and a kernel constraint with the max norm of 5. Then we use the dropout of 0.4 to handle the model over-fitting issues. The next layer consists of 256 neurons and a kernel constraint with the max norm of 5. The first two layers have RELU activations. Lastly, we have a single neuron layer with linear activation as we have BMI in the linear range. We employ Adam optimizer with default configurations and the loss function used is Mean Square Error (MSE) which is defined as:
\begin{eqnarray}
  \label{eq:modsoftmax}
  MSE = \frac{1}{n}\sum_{i=1}^{n}\left ( y_{i} - x_{i} \right )^2,
\end{eqnarray}
where $n$ is the number of samples, $y_{i}$ is the ground truth BMI, and $x_{i}$ is the predicted BMI value.

\section{Experiments}

\subsection{Dataset}
We used \textcolor{black}{three publicly available datasets} for evaluation of the proposed methodology: VisualBMI \cite{EnesKocabey} ,VIP attributes \cite{showme} and \textcolor{black}{Bollywood Dataset \cite{AiBmi}. We have not used FIW-BMI and Morph II
\footnote{\url{https://ebill.uncw.edu/C20231_ustores/web/store_main.jsp?STOREID=4}} because both datasets are not   available free of cost.}\\ 
\begin{figure*}[]
\centering
\includegraphics[width=\linewidth]{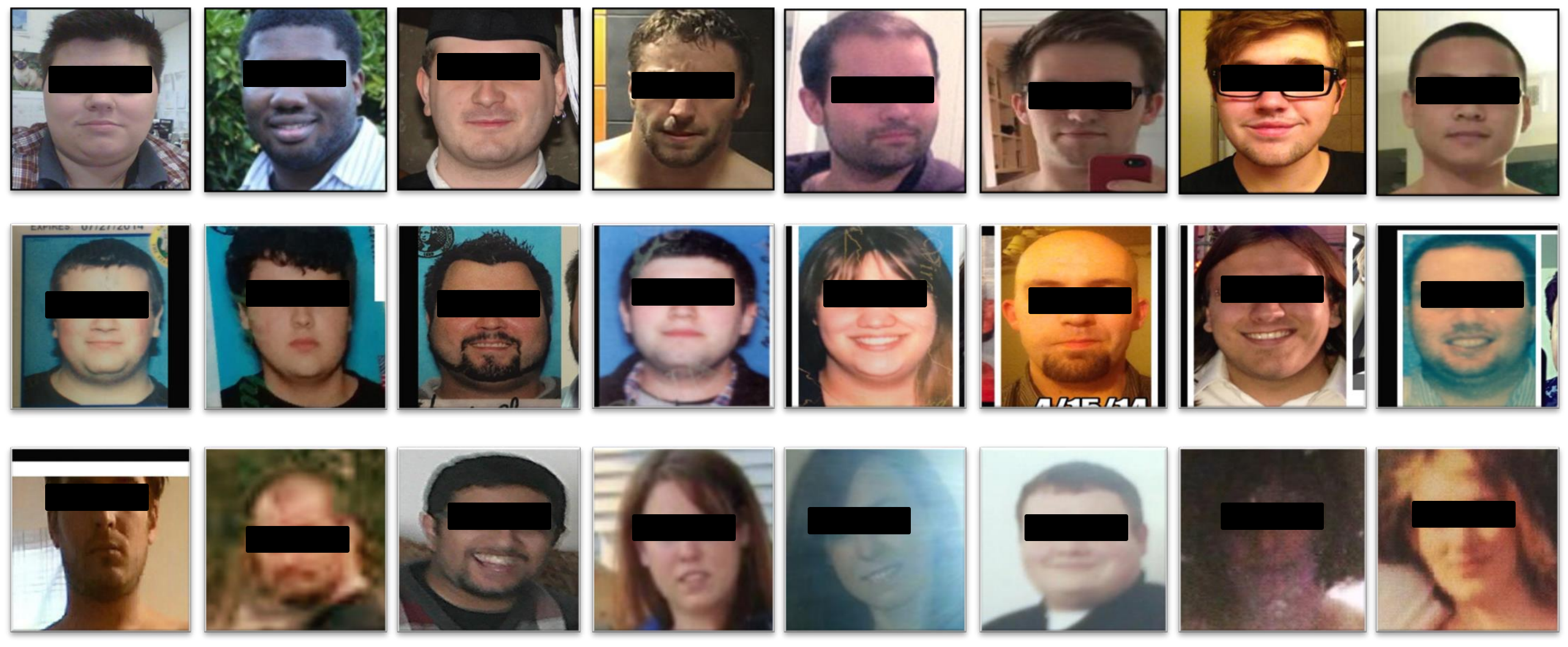}
\caption{Examples of face images in the VisualBMI dataset.}
\label{fig:VisualBMIdataset}
\end{figure*}

\noindent\textbf{VisualBMI}: The VisualBMI dataset is collected by \cite{EnesKocabey} from Reddit-subreddit called progress-pics where people share th-eir images of before and after body transformation. There are a total of 4206 images, out of which 2438 are males and 1768 are females. \textcolor{black}{We followed the split provided by original authors \cite{EnesKocabey}.} The first 3368 images are used for training and the rest of the images are used for testing. The dataset is quite challenging due to several low quality and variable size blurry images. Furthermore, some of the images are the images of the picture. The typical example of images are shown in Figure \ref{fig:VisualBMIdataset}.\\

\begin{figure*}[]
\centering
\includegraphics[width=\linewidth]{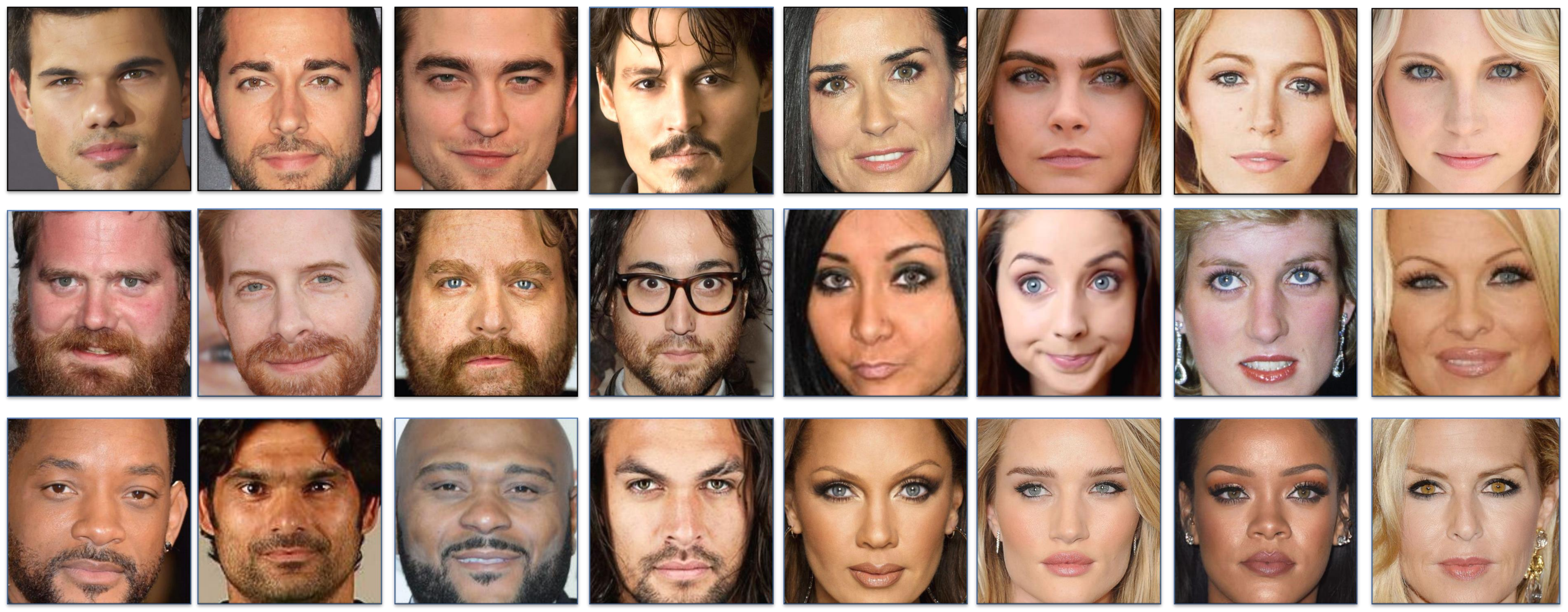}
\caption{Examples of face images in the VIP-attribute dataset.}
\label{fig:vip_dataset}
\end{figure*}

\noindent\textbf{VIP Attribute dataset}: The VIP attribute dataset is collected by Bilinski et al. \cite{showme}. This dataset consists of 1026 images of males (513) and females (513). \textcolor{black}{For a fair comparison, we have performed experiments using a 78/22 split of data which are provided by Jiang et al. \cite{labeldist}} The images are of singers and athletes. Unlike the VisualBMI dataset, these images are mainly frontal and are of high quality. This dataset is challenging due to the presence of makeup, plastic surgery, beard, and mustache. The authors of \cite{showme} collected the height and weight from different celebrity websites and calculated the BMI. The typical example of images is shown in Figure \ref{fig:vip_dataset}.\\

\noindent \textcolor{black}{\textbf{Bollywood dataset}: The `Bollywood' dataset is publicly available at GitHub \footnote{\url{https://bit.ly/2PNhDH7}}. This dataset contains 237 labeled images. All the images are of Bollywood celebrities. There are a total of 22 identities in this dataset which means that there are multiple images per celebrity. We used a 78/22 split for training and testing. The experimental results in Table \ref{Table:Bollywood_d} show the superiority of our approach on this new dataset as well.
}

\begin{table}[H]
\textcolor{black}{
\small\addtolength{\tabcolsep}{-3.2pt}
\centering
\begin{tabular}{|l|c|c|c|c|}
\hline
Model   & SVR \cite{AiBmi}& RR \cite{AiBmi} & Our (GAP) & Our  (Reg-GAP) \\ \hline
VGG19   & 1.99      & 1.49     & 0.98      & \textbf{0.55}  \\ \hline
VGGFace & 0.96      & 0.97     & 0.40      & \textbf{0.32}  \\ \hline
\end{tabular}
\caption{Results on the Bollywood dataset. Results show that Reg-GAP results are better than that of GAP and  \cite{AiBmi}.}
\label{Table:Bollywood_d}
}
\end{table}

\begin{table}[t]
\small\addtolength{\tabcolsep}{-5.7pt}
\begin{tabular}{|l|c|l|c|l|c|}
\hline
\phantom{aa}\textbf{Model}   & \phantom{aa}\textbf{$p$} & \phantom{aa}\textbf{Model} & \phantom{aa}\textbf{$p$} & \phantom{aa}\textbf{Model} & \phantom{aa}\textbf{$p$} \\ \hline
VGG16            & \phantom{22}0.45             & ResNet50       & \phantom{22}0.33             & ResNet50v2     & \phantom{22}0.37             \\ \hline
VGG19            & \phantom{22}0.42             & ResNet101      & \phantom{22}0.31             & ResNet101v2    & \phantom{22}0.39             \\ \hline
\textbf{VGGFace} & \phantom{22}\textbf{0.65}    & ResNet152      & \phantom{22}0.27    & ResNet152v2    & \phantom{22}0.4              \\ \hline
\textbf{FaceNet}          & \phantom{22}\textbf{0.61}             & MobileNet      & \phantom{22}0.48             & MobileNetV2    & \phantom{22}0.38             \\ \hline
DenseNet121      & \phantom{22}0.47             & DenseNet169    & \phantom{22}0.44             & DenseNet201    & \phantom{22}0.49             \\ \hline
\end{tabular}
\caption{Model Selection: To select the best pre-trained model for feature extraction,  we extract features from different pre-trained models and apply SVR on them to select the best model which gives the highest Pearson correlation for our problem.}
\label{model_selection}
\end{table}

\begin{table*}[H]
\centering
\begin{tabular}{|l|c|c|c|c|c|c|c|c|c|}
\hline
Model   & \multicolumn{3}{l|}{\phantom{222222}VisualBMI \cite{EnesKocabey}} & \multicolumn{3}{l|}{\phantom{222222222}GAP} & \multicolumn{3}{l|}{\phantom{2222222}Reg-GAP}                   \\ \hline
Metric  & MAE           & RMSE          & Pearson          & MAE   & RMSE  & Pearson  & MAE           & RMSE          & Pearson        \\ \hline
FaceNet & \textcolor{black}{5.38}             & \textcolor{black}{7.51}              & 0.61             & 5.23  & 7.04  & 0.645    & \textbf{5.03} & \textbf{6.92} & \textbf{0.663} \\ \hline
VGGFace & \textcolor{black}{5.16}              & \textcolor{black}{7.16}             & 0.65             & 5.22  & 7.03  & 0.644    & \textbf{4.99} & \textbf{6.94} & \textbf{0.659} \\ \hline
\end{tabular}
\caption{Results on the VisualBMI dataset using VGGFace and FaceNet models. Lower MAE and RMSE is better while the higher Pearson correlation is more useful. Results show that Reg-GAP results are better than that of GAP and \cite{EnesKocabey}. }
\label{VisualBMI_Completeresults_Person_combined}
\end{table*}

\subsection{Evaluation metrics}
The evaluation metrics used in this papers are mean square error (MSE), root mean square error (RMSE) and Pearson correlation. To make our paper self contained, we define each of evaluation metric below.
\begin{eqnarray}
  \label{eq:mae}
  MAE = (\frac{1}{n})\sum_{i=1}^{n}\left | y_{i} - \tilde{y_{i}} \right |
\end{eqnarray}
where $n$ is the number of samples, $y_{i}$ is the ground truth and $\tilde{y_{i}}$ is the predicted BMI.
\begin{eqnarray}
  \label{eq:rmse}
  %RMSE = \sqrt{\frac{1}{n}\Sigma_{i=1}^{n}{\Big(\frac{y_i -x_i}{\sigma_i}\Big)^2}}
  RMSE = \sqrt{\frac{1}{n}\sum_{i=1}^{n}\left ( y_{i} - \tilde{y_{i}} \right )^2}
\end{eqnarray}
where $n$ is the number of samples, $y_{i}$ is the ground truth and $\tilde{y_{i}}$ is the predicted BMI. Finally the Pearson correlation is defined as:
\begin{eqnarray}
  \label{eq:pearson}
  Pearson =  \frac{ \sum_{i=1}^{n}(x_i-\bar{x})(y_i-\bar{y}) }{%
        \sqrt{\sum_{i=1}^{n}(x_i-\bar{x})^2}\sqrt{\sum_{i=1}^{n}(y_i-\bar{y})^2}}
\end{eqnarray}
where $n$ is the number of samples, $x_i$ and $y_i$ are the individual's ground truth and predicted BMI for person $i$, $\bar{x}$ and $\bar{y}$ are the mean of ground truth and predicted BMIs. Lower MAE and RMSE and higher Pearson correlation represent the improved results.

\subsection{Face model selection}
Recently several deep convolutional neural networks-ba-sed recognition models have been proposed for high accuracy face recognition. We have experimented with several of those models and selected the one which performs better for our problem of face to BMI prediction. Table \ref{model_selection} shows the experimental results. For model selection, we extracted the features from the second last layer of each model and applied $\epsilon$-SVR on the features to predict BMI.
As can be seen that among several models, VGGFace \cite{vggface} and FaceNet\citep{facenet} have the highest Pearson correlation.
Therefore, for our experiments, we have chosen VGG-Face and FaceNet models. 

\subsection{Experimental results on VisualBMI dataset}
Table \ref{VisualBMI_Completeresults_Person_combined} shows the results of region-aware global pooling features on the VisualBMI dataset \cite{EnesKocabey} using VGGFace and FaceNet models. \textcolor{black}{Since the results using MAE and RMSE were not mentioned by the authors of \cite{EnesKocabey}, therefore, we have computed the results on these metrics using authors \cite{EnesKocabey} code.} The improved face to BMI predictions on all three evaluation metrics (MAE, RMSE, and Pearson correlation) demonstrate the usefulness of the proposed approach.
The superiority of Reg-GAP as compared to GAP shows that explicit feature pooling from different face regions is important and helps in better prediction of BMI from the face image.

\subsubsection{Class-level BMI prediction}

In this section, we present the BMI class division of the VisualBMI data set. We followed the division of \citep{EnesKocabey} and the details are shown in Table \ref{VisualBMI_classfication}. All samples under 18.5 are labeled underweight while all samples above 40 are labeled very severely obese.

Table \ref{Classlevel-BMI} shows the experimental results of BMI prediction for the different classes of datasets. Improved experimental results of Reg-GAP for all BMI-classes enforce our conjecture that, as compared to extracting only face-level features, face to BMI prediction methods should focus on various facial regions to get a better BMI prediction. Note that the `very severely obese' class has the highest MAE and RMSE due to the large class variation.

 \subsubsection{Gender prediction}
\begin{figure}[h!]
\centering
\includegraphics[width=\linewidth]{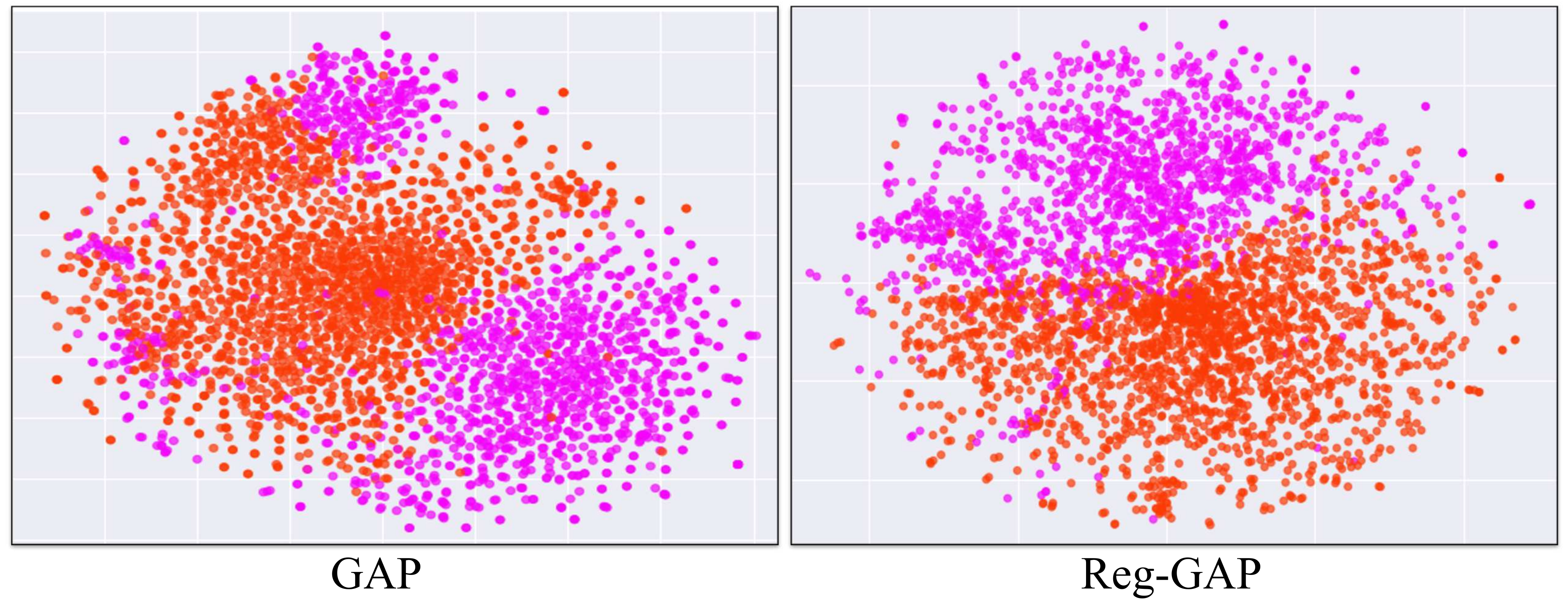}
\caption{t-SNE of the features of GAP and Reg-GAP for gender classes in VisualBMI dataset.}
\label{fig:tsne}
\end{figure}

To demonstrate the discriminative ability of the proposed Reg-GAP, we employ t-SNE to draw the features with GAP and Reg-GAP for the male and females. In Figure \ref{fig:tsne}, red dots show feature vectors for males, and pink dots show feature vectors for females, where the feature vectors are extracted from VGG-Face models for the VisualBMI dataset. The better separation of Reg-GAP features as compared to that of GAP features demonstrates the usefulness of Reg-GAP features.

 \newcolumntype{L}{>{\centering\arraybackslash}m{1cm}}
\begin{table}[]
\centering
\small\addtolength{\tabcolsep}{-1.7pt}
\begin{tabular}{|l|c|c|c|c|}
\hline
\multicolumn{1}{|c|}{Class} & \begin{tabular}[c]{@{}c@{}}Train\\ Images\end{tabular} & \begin{tabular}[c]{@{}c@{}}Test\\ Images\end{tabular} & BMI \textgreater{} & BMI \textless{} \\ \hline
Under Weight                & 7                                                      & 0                                                     & 16    & 18.5  \\ \hline
Normal                      & 555                                                    & 127                                                   & 18.5  & 25    \\ \hline
Over Weight                 & 936                                                    & 215                                                   & 25    & 30    \\ \hline
Obese                       & 772                                                    & 169                                                   & 30    & 35    \\ \hline
Severely Obese              & 541                                                    & 140                                                   & 35    & 40    \\ \hline
Very Severely Obese         & 557                                                    & 189                                                   & 40    & -     \\ \hline
\end{tabular}
\caption{BMI classes with train and test samples \textcolor{black}{for VisualBMI dataset.}}
\label{VisualBMI_classfication}
\end{table}

\begin{table}[]
\small\addtolength{\tabcolsep}{-4.3pt}
\centering
\begin{tabular}{|l|l|c|c|c|c|c|}
\hline
                      & \multicolumn{1}{c|}{Class} & Normal        & \begin{tabular}[c]{@{}c@{}}Over\\ Weight\end{tabular} & Obese         & \begin{tabular}[c]{@{}c@{}}Severely\\ Obese\end{tabular} & \begin{tabular}[c]{@{}c@{}}Very\\ Severely\\ Obese\end{tabular} \\ \hline
\multirow{2}{*}{MAE}  & GAP                        & 1.56          & 1.44                                                  & 1.80          & 1.72                                                     & 5.54                                                            \\ \cline{2-7} 
                      & Reg-GAP                    & \textbf{1.25} & \textbf{1.28}                                         & \textbf{1.22} & \textbf{1.23}                                            & \textbf{4.91}                                                   \\ \hline
\multirow{2}{*}{RMSE} & GAP                        & 1.97          & 1.78                                                  & 2.19          & 2.18                                                     & 8.08                                                            \\ \cline{2-7} 
                      & Reg-GAP                    & \textbf{1.56} & \textbf{1.52}                                         & \textbf{1.48} & \textbf{1.47}                                            & \textbf{7.69}                                                   \\ \hline
\end{tabular}
\caption{Comparisons of GAP with Reg-GAP for different BMI classes. Reg-GAP outpeforms GAP for all the classes \textcolor{black}{for VisualBMI dataset.}}
\label{Classlevel-BMI}
\end{table}

\begin{table}[t]
\small\addtolength{\tabcolsep}{-4.7pt}
\begin{tabular}{|l|l|l|l|l|l|}
\hline
     All   & \begin{tabular}[c]{@{}l@{}}ResNet\\Based\citep{showme}\end{tabular} & LD-PLS\citep{labeldist} & LD-CCA\citep{labeldist} &  GAP & Reg-GAP \\ \hline
MAE     & \phantom{22}2.36           & \phantom{22}2.26   & \phantom{22}2.23   & \phantom{2}\textcolor{black}{1.85}     & \phantom{22}\textcolor{black}{\textbf{1.73}}     \\ \hline
RMSE    & \phantom{22}\phantom{22}-              & \phantom{22}\phantom{22}-      & \phantom{22}\phantom{22}-      & \phantom{2}\textcolor{black}{2.74}     & \phantom{22}\textcolor{black}{\textbf{2.61}}     \\ \hline
Pearson & \phantom{22}0.55              & \phantom{22}\phantom{22}-      & \phantom{22}\phantom{22}-      & \phantom{2}\textcolor{black}{0.71}    & \phantom{22}\textcolor{black}{\textbf{0.75}}    \\ \hline
\end{tabular}
\caption{Results on the entire VIP attribute dataset. Lower MAE/RMSE and higher Pearson correlation are better. Results show that Reg-GAP results are better than that of GAP and \cite{labeldist}. }
\label{Table:Entrire_BMI_Dataset}
\end{table}

\subsection{Experimental results on VIP-Attribute Dataset}
In this section, we show the results of the proposed approach on the VIP-attributes dataset and compare our method with the recent state-of-the-art approach. Table \ref{Table:Entrire_BMI_Dataset} shows our results for the entire VIP attribute dataset using the VGGFace model.  
Table \ref{Table:female_BMI_Dataset} and Table \ref{Table:Male_BMI_Dataset} shows the experimental results for males and females separately for the VIP-attribute dataset. The overall and gender-wise experimental results demonstrate the usefulness of the proposed approach.

\begin{table}[t]
\small\addtolength{\tabcolsep}{-4.7pt}
\begin{tabular}{|l|l|l|l|l|l|}
\hline
     Female   & \begin{tabular}[c]{@{}l@{}}ResNet\\Based\citep{showme}\end{tabular}  & LD-PLS\citep{labeldist} & LD-CCA\citep{labeldist} & GAP & Reg-GAP \\ \hline
MAE     & \phantom{22}2.30           & \phantom{22}2.28   & \phantom{22}2.27   & \phantom{2}\textcolor{black}{1.80}    & \phantom{22}\textbf{\textcolor{black}{1.63}}     \\ \hline
RMSE    & \phantom{22}\phantom{22}-              & \phantom{22}\phantom{22}-      &\phantom{22}\phantom{22}-      & \phantom{2}\textcolor{black}{2.84}     & \phantom{22}\textbf{\textcolor{black}{2.53}}     \\ \hline
Pearson &   \phantom{22}{0.55}            & \phantom{22}\phantom{22}-      & \phantom{22}\phantom{22}-      & \phantom{2}\textcolor{black}{0.74}     & \phantom{22}\textbf{\textcolor{black}{0.81}}     \\ \hline
\end{tabular}
\caption{Results on the female class of VIP attribute dataset.  Results show that Reg-GAP results are better than that of GAP and \cite{labeldist}.}
\label{Table:female_BMI_Dataset}
\end{table}

\begin{table}[t]
\small\addtolength{\tabcolsep}{-4.7pt}
\begin{tabular}{|l|l|l|l|l|l|}
\hline
     Male & \begin{tabular}[c]{@{}l@{}}ResNet\\Based\citep{showme}\end{tabular}& LD-PLS\citep{labeldist} & LD-CCA\citep{labeldist} & GAP & Reg-GAP \\ \hline
MAE     & \phantom{22}2.32           & \phantom{22}2.25   & \phantom{22}2.19   & \phantom{2}\textcolor{black}{1.89}     & \phantom{22}\textbf{\textcolor{black}{1.77}}     \\ \hline
RMSE    & \phantom{22}\phantom{22}-              & \phantom{22}\phantom{22}-      & \phantom{22}\phantom{22}-      & \phantom{2}\textcolor{black}{2.71}     & \phantom{22}\textbf{\textcolor{black}{2.66}}     \\ \hline
Pearson & \phantom{22}\textbf{0.55}              & \phantom{22}\phantom{22}-      & \phantom{22}\phantom{22}-      & \phantom{2}\textcolor{black}{0.37}     & \phantom{22}\textcolor{black}{0.41}     \\ \hline
\end{tabular}
\caption{Results on the male class of VIP-attribute dataset. Results show that Reg-GAP results are better than that of GAP and \cite{labeldist}.}
\label{Table:Male_BMI_Dataset}
\end{table}

\begin{table*}
\textcolor{black}{
\centering
\begin{tabular}{|l|c|c|c|c|c|c|c|c|} 
\hline
\multirow{2}{*}{Dataset} & \multicolumn{3}{l|}{VIP\_Attribute\cite{showme}} & \multicolumn{2}{l|}{VisualBMI\cite{EnesKocabey}} & \multicolumn{2}{l|}{Bollywood\cite{AiBmi}} & \multirow{2}{*}{P\_Value}                                                                        \\ 
\cline{2-8}
                         & Overall & Male & Female             & FaceNet & VGGFace              & VGG19 & VGGFace                &                                                                                                  \\ 
\hline
Recent Work              & 2.23    & 2.19 & 2.27               & 5.38    & 5.16                 & 1.49  & 0.97                   & \multicolumn{1}{c|}{\multirow{2}{*}{\begin{tabular}[c]{@{}c@{}}\\0.001392827177 \end{tabular}}}  \\ 
\cline{1-8}
Ours                     & 1.73    & 1.77 & 1.63               & 5.03    & 4.99                 & 0.55  & 0.32                   & \multicolumn{1}{c|}{}                                                                            \\
\hline
\end{tabular}
\caption{Results of significance tests on MAE of all three datasets.}
\label{Table:significance_test}
}
\end{table*}

\textcolor{black}{
\subsection{\textcolor{black}{Run-Time Analysis}}
In our experiments, we used Google Colab with RAM of 25GB and ROM of 68GB, with the Tensor-flow version 1.14.0, Keras version 2.2, and Python 3.0. The optimizer used is Adam with learning rate of 0.001, and beta1=0.9, beta2=0.999, epsilon=0.48, decay=0.0. For each image, on average, the MTCNN module took 0.76s to detect the faces, the BiseNet took 0.07s for face semantic segmentation, and later reprocessing took 0.14s. The VGGFace module took 0.27s for generating Reg-GAP features and lastly the regression module on average took 2.63E-04s. The entire process of face to BMI prediction for each image took 1.24s.
}

\section{Discussion and Concluding remarks}
Determining BMI from facial photos is commonly used in recent studies where all of the previous methods have focused on the overall faces but we, on the other hand, have specifically tried to pool features from different face regions. To achieve accurate and pixel-wise localization, we employed face semantic segmentation. Our experimental results show that face to BMI prediction is improved with Reg-GAP as compared to using GAP. The graph in Figure \ref{fig:tsne} shows that, in addition to the face to BMI prediction problem, Reg-GAP features are more discriminative for the gender prediction problem. To the best of our knowledge, this is one of the first frameworks to utilize facial regions for the BMI prediction while maintaining the state of art accuracy. We have performed experiments on three publicly available datasets and comparisons over several evaluation metrics to validate the proposed ideas. \textcolor{black}{ Table \ref{Table:significance_test} shows the result of the significance test on all three datasets with the p-value of 0.0014. It indicates that our results are significant. We have also tested the normality of our samples with the Shapiro–Wilk test and it was also passed with the p-value of 0.0170625}

There are still some limitations in our framework. For example, semantic segmentation and feature extraction are being performed separately. Future work could include training of feature extractors and semantic segmentation in a joint framework. For instance, future work may use some student-teacher approach where training is done through the teacher model on semantically segmented facial images and the student model is enforced to predict the same embedding for the same input sample as the teacher model. This approach can let the student model behave as if it was using the segmentation without actually using the semantic segmentation. 

The second limitation for the BMI prediction problem is the biases in the BMI-datasets. Mostly the underweight class and very severely obese class are biased because there are very few samples in the underweight class and the very severely obese class has a large range. This can be addressed using losses that handle class imbalance such as focal loss. Lastly, since the datasets are collected from social media or the world wide web, there may be some errors in the annotations. The VisualBMI dataset has many poor quality images as some of the images are images of the pictures and some are blurry. Although the VIP attribute dataset has images in good quality, however, these images are of actors and there is the presence of makeup, plastic surgery which can make the accurate prediction of BMI difficult.\newline\newline

\section{Appendix}
\noindent\textbf{Declaration of Competing Interest:} We wish to confirm that there are no known conflicts of interest associated with
this publication and there has been no significant financial support for this work that
could have influenced its outcome. We confirm that the manuscript has been read and
approved by all named authors and that there are no other persons who satisfied the
criteria for authorship but are not listed. We further confirm that the order of authors
listed in the manuscript has been approved by all of us. We confirm that we have given
due consideration to the protection of intellectual property associated with this work
and that there are no impediments to publication, including the timing of publication,
with respect to intellectual property. In so doing we confirm that we have followed the
regulations of our institutions concerning intellectual property. We understand that the
Corresponding Author is the sole contact for the Editorial process (including Editorial
Manager and direct communications with the office). He/she is responsible for communicating
with the other authors about progress, submissions of revisions, and final
approval of proofs. We confirm that we have provided a current, correct email address
which is accessible by the Corresponding Author.
\newline\newline

%\noindent\textbf{Declaration of Competing Interest:} The authors declare that they have no known competing financial interests or personal relationships that could have appeared to influence the work reported in this paper. The authors also confirm that the manuscript has been read and approved by all named authors and that there are no other persons who satisfied the criteria for authorship but are not listed and the order of authors listed in the manuscript has been approved by all of the authors.\newline\newline
\noindent\textbf{Credit authorship contribution statement:}\newline
\textbf{Nadeem Yousaf:} conceptualization, software, writing - original draft, writing-review \& editing \textbf{Sarfaraz Hussein:} conceptualization, writing-review, and editing, idea, formal analysis, supervision, project administration. \textbf{Waqas Sultani:} conceptualization, writing-review, and editing, idea, formal analysis, supervision, project administration.

%% Loading bibliography style file
%\bibliographystyle{model1-num-names}
\bibliographystyle{cas-model2-names}

% Loading bibliography database
\bibliography{cas-refs}

%\vskip3pt

\end{document}